\documentclass{article}

\usepackage[nonatbib,final]{nips_2017}

\usepackage[utf8]{inputenc} 
\usepackage[T1]{fontenc}    
\usepackage{hyperref}       
\usepackage{url}            
\usepackage{booktabs}       
\usepackage{amsfonts}       
\usepackage{nicefrac}       
\usepackage{microtype}      

\usepackage{times}

\usepackage{amssymb, amsmath}
\usepackage{mathtools}
\usepackage{graphicx}
\usepackage{subfigure}
\usepackage{enumitem}
\usepackage{url}            
\usepackage{booktabs}       
\usepackage{amsfonts}       
\usepackage{nicefrac}       
\usepackage{microtype}      
\usepackage{color}
\usepackage{algorithm}
\usepackage[noend]{algpseudocode}

\makeatletter
\def\BState{\State\hskip-\ALG@thistlm}
\makeatother

\newcommand*{\x}{\mathbf{x}}
\newcommand*{\xk}{\mathbf{x}_k}
\newcommand*{\xkk}{\mathbf{x}_{k+1}}
\newcommand*{\uu}{\mathbf{u}}
\newcommand*{\uk}{\mathbf{u}_k}
\newcommand*{\huk}{\hat{\mathbf{u}}_k}
\newcommand*{\dt}{\Delta t}
\newcommand*{\qk}{q_k\dt}
\newcommand*{\E}{\mathbb{E} }
\newcommand*{\xx}{\x_{k+1}|\xk }
\newcommand*{\xu}{\xk,\uk }
\newcommand*{\xxu}{\x_{k+1}|\xu }

\newcommand*{\Vavg}{V_{avg} }
\newcommand*{\hVavg}{\hat{V}_{avg}}

\newcommand*{\expq}{e^{-q_k\dt} }

\newcommand*{\R}{S}
\newcommand*{\hR}{\hat{S}}
\newcommand*{\Rinv}{\R^{-1} }

\newcommand*{\Zk}{Z_k }
\newcommand*{\Zkk}{Z_{k+1} }
\newcommand*{\Zkt}{\hat{Z}_{k} }
\newcommand*{\Zkkt}{\hat{Z}_{k+1} }
\newcommand*{\Zavg}{Z_{avg}}
\newcommand*{\hZavg}{\hat{Z}_{avg}}

\newcommand*{\paramz}{\boldsymbol{\nu} }

\newcommand*{\Vkt}{\hat{V}_{k} }

\newcommand*{\trim}[1]{}

\newcommand*{\comm}[1]{}
\def\diag{\mathop{\rm diag}\nolimits}

\title{Actor-Critic for Linearly-Solvable Continuous MDP\\ with Partially Known Dynamics}

\author{
  Tomoki Nishi \\  Toyota Central R \& D Labs. Inc,\\ Nagakute, Aichi, Japan\\ \texttt{nishi@mosk.tytlabs.co.jp} \\
 \And
 Prashant Doshi \\ THINC Lab, Dept. of Computer Science \\ University of Georgia, Athens, GA 30622\\ \texttt{pdoshi@cs.uga.edu}
 \And Michael R. James \\Toyota Research Institute \\ Ann Arbor, MI, 48105 \\ \texttt{michael.james@tri.global}
 \And Danil Prokhorov \\Toyota R \& D\\ Ann Arbor, MI, 48105\\  \texttt{danil.prokhorov@toyota.com}
}

\begin{document}

\maketitle

\begin{abstract}
 In many robotic applications, some aspects of the system dynamics can be modeled accurately while others are difficult to obtain or model. We present a novel reinforcement learning (RL) method for continuous state and action spaces that learns with partial knowledge of the system and without active exploration. It solves linearly-solvable Markov decision processes (L-MDPs), which are well suited for continuous state and action spaces, based on an {\em actor-critic} architecture. Compared to previous RL methods for L-MDPs and path integral methods which are model based, the actor-critic learning does not need a model of the uncontrolled dynamics and, importantly, transition noise levels; however, it requires knowing the control dynamics for the problem. We evaluate our method on two synthetic test problems, and one real-world problem in simulation and using real traffic data. Our experiments demonstrate improved learning and policy performance.
\end{abstract}

\section{Introduction}
\label{sec:intro}
Reinforcement learning (RL) offers a way of learning high-quality policies (control) for an agent by exploring its environment. Methods for RL have predominantly focused on domains with discrete state and actions. Those that operate on continuous states or actions resort to sampling or other approximations because of the difficulty in analytically solving the continuous Bellman equation~\cite{wiering2012reinforcement}. In this regard, Todorov~\cite{todorov2006linearly} introduced the {\em linearly-solvable Markov decision process} (L-MDP), a subclass of general MDPs, which allows us to quickly solve the continuous Bellman equation exactly under a class of structured dynamics and rewards. Specifically, the Bellman equation in L-MDPs is recast as a linearized differential, and its solution is efficiently obtained as a linear eigenfunction when the whole dynamics model is available~\cite{todorov2009eigenfunction}. As such, L-MDPs are particularly well suited for modeling robotic learning and planning, where the state and actions spaces are usually continuous.

In addition to continuous spaces, high-impact robotic applications such as autonomous vehicles impose an additional constraint on RL. They preclude an exhaustive exploration of the state and action space because it would be unacceptable for an autonomous vehicle to optimistically try maneuvers that would lead to a crash or even a near-miss, leaving much of the state and action spaces unexplored. However, at the same time, guaranteeing safe exploration has a high computational cost that is shown to be NP-hard~\cite{moldovan2012safe}.

L-MDPs decompose the dynamics model into {\em passive} and {\em active (control)} dynamics with added actuator noise. In this paper, we present a new method for semi model-free RL for L-MDPs, which uses a partially-known system dynamics model. Specifically, the method requires the control dynamics model, which represents the effect of actions to be specified, but not the passive dynamics model nor the noise in the transitions. Knowing control dynamics is feasible in our motivating context of autonomous driving because advanced driver assistance systems such as adaptive cruise control are already available in most new vehicles, and these utilize a model of the control dynamics. More importantly, knowing such a model makes safe active exploration unnecessary. Furthermore, the correct actuator noise is often difficult to prespecify, which adds to this method's appeal.

Our method called {\em passive actor-critic} (pAC) finds the policy within the standard two-step architecture of actor-critic methods comprising a policy improvement step (actor) and state evaluation (critic). Passive actor-critic combines the data collected on passive state transitions with the known control dynamics. In the context of L-MDPs, the critic estimates the expected value function using the linearized Bellman equation from the data and the actor improves the policy using the standard Bellman equation on data and the active dynamics model. In addition to the well-known radial basis function~\cite{todorov2009eigenfunction,uchibe2014combining}, multi-layer neural networks are also introduced for approximating the value function with demonstrated performance improvements. The method is evaluated on two known synthetic domains and on our motivating domain of freeway merge by an autonomous car both in simulation and using real-world traffic data. Interestingly, pAC improves on previous model-based methods despite requiring reduced model specifications. Importantly, pAC finds policies that succeed in freeway merge on real data at rates exceeding 90\%, motivating transition to real-world testing as future work.

\section{Related Work}
\label{sec:related}

Previous approaches for solving L-MDPs are predominantly  model based~\cite{todorov2009efficient,todorov2009eigenfunction,zhong2011aggregation}. These efficiently optimize control policies by solving the linearized Bellman in discrete- or continuous-state L-MDPs when the system dynamics is fully known. Our method relaxes this requirement using samples of passive dynamics, while knowing the control dynamics. We also introduce multi-layer neural networks for approximating the value functions in L-MDPs in addition to the previously used radial basis functions.

Uchibe and Doya~\cite{uchibe2014combining} formulate Z-learning based on least-square TD learning for continuous L-MDPs. The method optimizes the policy while requiring knowledge of the control dynamics and transition noise. The motion noise inherent in robotic platforms is often unknown due to which this method may not apply to robot learning. pAC learns the noise levels during optimization given sampled data and knowledge of control dynamics by minimizing the error between the value and action-value functions. This makes pAC well positioned for application to robot-based RL.

As pAC can learn from data containing samples of passive dynamics, it bears resemblance to batch RL methods~\cite{lange2012batch}. A popular and model-free batch RL method is fitted Q-iteration~\cite{antos2008fitted,gaeta2016fitted}, which finds policy from collected data without a model of system dynamics. It searches for actions that minimize the Q-value, which requires that either the action space be discrete or the Q-function has structure such as being quadratic due to computational cost. In contrast, pAC uses policy that is analytically derived from the estimated Z-value, parameter for transition noise, and known control dynamics.

Path integral control also learns a policy based on linearized Bellman equation~\cite{Theodorou2010,Gomez2014}. Unlike approaches for L-MDPs, path integral control can directly optimize the policy. However, the approach has to sample sample many trajectories under a training policy from a certain initial state. As we mentioned previously, we seek to avoid such active and potentially unsafe explorations in the real world.

Finally, many RL methods that use neural networks for continuous MDPs have been recently  proposed~\cite{heess2015learning,schulman2015trust}. These methods optimize a policy with active exploration while our method seeks to find a policy from data on passive state transitions (of the underlying Markov chain) with known control dynamics. On the other hand, the utility of networks to approximate the value function remains the same.

\section{Preliminaries}
\label{sec:preliminaries}

We briefly review L-MDPs for understanding our method.
%
We focus on a discrete-time system with a real-valued state $\x \in \mathbf{R}^n $ and control input $\mathbf{u} \in \mathbf{R}^m$, whose stochastic dynamics is defined as follows:
\vskip -7mm
\begin{align}
 \xkk  & = \xk + A(\xk)\dt + B\uk\dt + \diag\left(\boldsymbol{\sigma}\right)\boldsymbol{\Delta\omega}, \label{eq:dynamics}
\end{align}
\vskip -4mm
where $\boldsymbol{\Delta\omega}$ is differential Brownian motion simulated by a Gaussian $\mathcal{N}(\mathbf{0},\mathbf{I}\dt)$, where $\mathbf{I}$ is the identity matrix. $A(\xk)$, $B\mathbf{\uk}$ and $\boldsymbol{\sigma} \in \mathbf{R}^n$ denote the {\em passive} dynamics, {\em control} dynamics due to action, and the transition noise level, respectively ($B$ is an input-gain matrix). $\Delta t$ is a step size of time and $k$ denotes a time index. System dynamics structured in this way are quite general: for example, models of many mechanical systems conform to these dynamics.


L-MDP~\cite{todorov2009efficient} is a subclass of MDPs~\cite{puterman1994Markov} defined by a tuple, $\langle \mathcal{X},\mathcal{U},\mathcal{P},\mathcal{R}\rangle$, where $\mathcal{X} \subseteq \mathbf{R}^n$ and $\mathcal{U} \subseteq \mathbf{R}^m$ are continuous state and action spaces. $\mathcal{P}\coloneqq \{p(\mathbf{y}|\mathbf{x},\mathbf{u}) \>|\> \mathbf{x},\mathbf{y} \in \mathcal{X}, \mathbf{u}\in \mathcal{U} \}$ is a state transition model due to action, which is structured as in Eq.~\ref{eq:dynamics}, and $\mathcal{R} \coloneqq \{r(\mathbf{x},\mathbf{u})\>|\> \mathbf{x}\in \mathcal{X}, \mathbf{u}\in \mathcal{U}\}$ is an immediate cost function with respect to state $\mathbf{x}$ and action $\mathbf{u}$. A control policy $\mathbf{u}=\pi(\mathbf{x})$ is a function that maps a state $\mathbf x$ to an action $\mathbf u$. The goal is to find a policy that minimizes the following average expected cost: $\Vavg \coloneqq \lim_{n\rightarrow\infty}\frac{1}{n}\E\left[\sum_{k=0}^{n-1}r(\xk,\pi(\xk))\right]$.

Grondman et al.~\cite{grondman2012survey} notes that the Bellman equation for MDPs can be rewritten using the value function $V(\x)$ called {\em V-value}, state-action value function $Q(\x,\mathbf{u})$ called {\em Q-value}, and average value $V_{avg}$ under an policy.
\vskip -7mm
\begin{align}
\Vavg + Q_{k} &= r_{k} + \E_{p(\xkk|\xu)} [ V_{k+1}].
\label{eq:Bellman_rev}
\end{align}
\vskip -3mm
As we may expect, $V_{k} = \min_{\uu\in\mathcal{U}}Q_{k}$. $\E_{p(\xkk|\xk)}[\cdot]$ is expectation over a probability distribution of state transition under the passive dynamics. Here and elsewhere, subscript $k$ is values at time step $k$.

An L-MDP defines the cost of an action (control cost) to be the amount of stochastic effect it has on the system, adding it to the state cost:
\vskip -6mm
\begin{align}
r(\xu) \coloneqq q(\xk)\dt + KL(p(\xx)||p(\xxu)).
\label{eq:totalcost}
\end{align}
\vskip -2mm
Here, $q(\mathbf{x})\geq 0 $ is the state-cost function; $KL(\cdot||\cdot)$ is the Kullback-Leibler (KL) divergence; $p(\xx)$ models the {\em passive} dynamics while $p(\xxu)$ represents the {\em active} or control dynamics of the system. L-MDPs further add a condition on the dynamics as shown below.
\vskip -6mm
\begin{align*}
p(\xx) = 0 \Rightarrow \forall\uk~ p(\xxu) = 0.
\end{align*}
\vskip -2mm
This condition ensures that no action introduces new transitions that are not achievable under passive dynamics.\trim{In other words, actions are seen as simply contributing to the passive dynamics.} The stochastic dynamical system represented by Eq.~\ref{eq:dynamics} satisfies this assumption naturally because the dynamic is Gaussian. However, systems that are deterministic under passive dynamics remain so under active dynamics. This condition is easily met in robotic systems where noise is prevalent.

The standard Bellman equation for MDPs can then be recast in L-MDPs to be a linearized differential equation for exponentially transformed value function of Eq.~\ref{eq:Bellman_rev} (hereafter referred to as the linearized Bellman equation)~\cite{todorov2009eigenfunction}:
\vskip -6mm
\begin{align}
\Zavg\Zk & = \expq~\E_{p(\xkk|\xk)}[\Zkk] \label{eq:definition-of-z},
\end{align}
\vskip -2mm
where $\Zk \coloneqq e^{-V_k}$ and  $\Zavg \coloneqq e^{-\Vavg}$. Here, $\Zk$ and $Z_{avg}$ are an exponentially transformed value function called {\em Z-value} and the average cost under an optimal policy, respectively. Because the passive and control dynamics with the Brownian noise are Gaussian, the KL divergence between these dynamics becomes,
\vskip -6mm
\begin{align}
KL(p(\xx)&||p(\xxu)) = 0.5\uk ^\top \Rinv \uk \dt,
\label{eq:rho_}
\end{align}
\vskip -3mm
where $\Rinv \coloneqq B^\top ( \diag(\sigma _i^2) )^{-1}B$ and $\sigma _i$ denotes {\em i}-th element of $\boldsymbol{\sigma}$. Then, the optimal control policy for L-MDPs can we derived as,
\vskip -6mm
\begin{align}
\pi(\xk) & = -\R B^\top \frac{\partial V_k}{\partial \xk}.
\label{eq:optimraw}
\end{align}
\vskip -7mm

\section{Passive Actor-Critic for L-MDP}
\label{sec:pAC}
\trim{
\begin{figure}[!t]
\centering
\includegraphics[width=0.5\hsize]{overview-pAC.png}
\caption{\small Visualization of steps in pAC. {\em Critic} estimates the Z-value and the average cost using the linearized Bellman equation from samples under passive dynamics, and the {\em actor} improves on the Z-value using the known control dynamics, the Z-value and cost from the critic.}
\label{fig:overview}
\end{figure}
}
We present a novel actor-critic method for continuous L-MDP, which we label as {\em passive actor-critic} (pAC). \trim{Figure~\ref{fig:overview} shows a detailed schematic of pAC.} While the actor-critic method usually operates using samples collected actively in the environment~\cite{konda1999actor}, pAC finds a converged policy without exploration. Instead, it uses samples of passive state transitions and a known control dynamics model. pAC follows the usual two-step schema of actor-critic: a state evaluation step (critic), and a policy improvement step (actor).
\begin{enumerate}[leftmargin=*, itemsep=0pt, topsep=0pt]
\item {\bf Critic}: Estimate the Z-value and the average cost from the linearized Bellman equation using samples under passive dynamics;
\item {\bf Actor}: Improve a control policy by optimizing the Bellman equation given the known control dynamics model, and the Z-value and cost from the critic.
\end{enumerate}
We provide details on these two components below.

\subsection{Estimation by Critic using Linearized Bellman}

The critic step of pAC estimates Z-value and the average cost by minimizing the least-square error between the true Z-value and estimated one denoted by $\hat{Z}$.
\vskip -7mm
\begin{align}
\min_{\paramz,\hZavg}\frac{1}{2}\int _{\x}\Big( \hZavg\hat{Z}(\x;\paramz) -& \Zavg Z(\x)\Big)^2 d\x, \label{eq:ls_error}\\
 {\bf s.t.} \, \int _{\x} \hat{Z}(\x;\paramz) d\x = C,&\,\, \forall {\mathbf x} \,\, 0 < \hat{Z}(\x;\paramz) \leq  \frac{1}{\hZavg}, \nonumber
 \end{align}
\vskip -3mm
where $\paramz$ is a parameter vector of the approximation and $C$ is a constant value used to avoid convergence to the trivial solution $\hat{Z}(\x;\paramz)= 0$ for all $\x$. The second constraint comes from $\forall {\x},\, Z(\x) \coloneqq e^{-V(\x)} > 0$ and $\forall {\mathbf x},\, q({\mathbf x})\geq 0$. The latter implies that $V + \Vavg > 0$, and note that $\Zavg Z(x) := e^{-(V+ \Vavg)}$, which is less than 1.

We minimize the least-square error in Eq.~\ref{eq:ls_error}, $\hZavg\Zkt - \Zavg\Zk$, with TD-learning. The latter minimizes TD error instead of the least-square error that requires the true $Z(\x)$ and $\Zavg$, which are not available. The TD error denoted as $e^{i}_k$ for linearized Bellman equation is defined using a sample $(\xk,\xkk)$ of passive dynamics as, $e^{i}_k \coloneqq \hZavg^{i}\Zkt^{i} - e^{-q_k}\Zkkt^{i}$, where the superscript $i$ denotes the iteration.
$\hZavg$ here is updated using the gradient as follows:
\vskip -7mm
\begin{align}
& \hZavg^{i+1} = \hZavg^{i} - \alpha_1^{i}\frac{\partial \left ( e^{i}_k \right)^2}{\partial \hZavg} = \hZavg^{i} - 2\alpha_1^{i} e^{i}_k \Zkt^{i},\label{eq:update_zavg}
\end{align}
\vskip -4mm
where $\alpha^{i}_1$ is the learning rate, which may adapt with iterations.

In this work, we approximate the Z-value function in two ways: $(i)$ using a linear combination of weighted RBFs, and $(ii)$ using a neural network (NN). When a NN with an exponentiated activation function of output layer is used, the parameters $\paramz$ are updated with the following gradient based on backpropagation.~\footnote{$e^{-\rm{tanh}(x)}$ or $e^{-\rm{softplus}(x)}$ is used as an activation function of the output layer to satisfy the constraint $\hat{Z}\geq 0$. The constraint $\int _{\x} \hat{Z}(\x;\paramz) d\x = C$  is ignored in practice because convergence to  $\forall{\x},\, \hat{Z}(\x;\paramz)=0$ is rare. $\min{([1,e^{-q_k}\Zkkt^{i}])}$ is used instead of $e^{-q_k}\Zkkt^{i}$ to satisfy $\hat{Z} \leq 1/\Zavg$ in Eq.~\ref{eq:ls_error}.}
\vskip -6mm
\begin{align}
& \frac{\partial}{\partial \paramz^{i}}\left (\hZavg\Zkt^{i} - \Zavg\Zk  \right)^2 \approx 2e^{i}_k \hZavg^i\frac{\partial \Zkt^{i}}{\partial \paramz^{i}}, \label{eq:update_omega}
\end{align}
\vskip -4mm
where $e_k^i$ is the TD error as defined previously.
\trim{$\nabla_{\paramz^{i}}\Zkkt^i$ is a partial derivative operator with respect to $\paramz^{i}$ on $\Zkkt^i$.}

On the other hand, when weighted RBFs are used, $\hat{Z}(\x;\paramz)\coloneqq \paramz^{\top}\mathbf{f}(\x)$, and a Lagrangian relaxation of the objective function is useful as it includes the three constraints weighted using Lagrangian parameters $\lambda_1$, $\boldsymbol{\lambda}_{2}$ and $\lambda_3$. For convenience, denote $\tilde{\paramz}^{i}$ as,
\vskip -6mm
\begin{align*}
\tilde{\paramz}^{i} &\coloneqq \paramz^{i} - 2\alpha_2^{i} e^{i}_k\hZavg^{i}\frac{\partial \Zkt^{i}}{\partial \paramz^{i}},
\end{align*}
\vskip -4mm
where $\alpha_2^{i}$ is the learning rate. The update of $\paramz$ cognizant of the constraints is then,
\begin{align}
  \paramz^{i+1} &=  \tilde{\paramz}^{i} +\frac{\partial}{\partial \tilde{\paramz}^{i}}\left(\lambda^{i}_1\left(\int _{\x} \tilde{\paramz}^{i\top}\mathbf{f}(\x) d\x - C\right) + \boldsymbol{\lambda}_{2}^{i\top}\left(\tilde{\paramz}^{i}-\mathbf{0}\right)+ \lambda^{i}_3\left(\hat{Z}_k-1/\hZavg\right) \right),\nonumber\\
 &= \tilde{\paramz}^{i} + \lambda^{i}_1\mathbf{1}+ \boldsymbol{\lambda}^{i}_2+\lambda^{i}_3\mathbf{f}_k, \label{eq:omega_updated_constraint}
\end{align}
where we utilize $\int _{\x} \mathbf{f(\x)}d\x = \mathbf{1}$ and replace the constraint $\forall {\mathbf x} \,\, \hat{Z}(\x;\paramz) > 0$ by $\paramz > \mathbf{0}$ because the constraint on $\paramz$ always satisfies the former. $\mathbf{1}$ is a vector of all ones.

Lagrangian parameters $\lambda_1$, $\boldsymbol{\lambda}_2$ and $\lambda_3$ in each iteration are obtained by ensuring that the updated parameter vector $\paramz^{i+1}$ satisfies the three constraints in Eq. ~\ref{eq:ls_error} for $\xk$. Formally,
\vskip -7mm
\begin{align*}
 &\hat{Z}(\xk;\paramz^{i+1}) = ( \tilde{\paramz}^{i} + \lambda^{i}_1\mathbf{1} +\boldsymbol{\lambda}^{i}_{2}+\lambda^{i}_3 \mathbf{f}_k )^{\top}\mathbf{f}_k \leq 1/\hZavg,\\
 &\paramz^{i+1} = \tilde{\paramz}^{i} + \lambda^{i}_{1}\mathbf{1} + \boldsymbol{\lambda}^{i}_{2} + \lambda^{i}_{3}\mathbf{f}_{k} > \mathbf{0}~ \text{and},\\
 &\int _{\x} \hat{Z}(\x;\paramz^{i+1}) d\x  = \left( \tilde{\paramz}^{i} + \lambda^{i}_1\mathbf{1} +\boldsymbol{\lambda}^{i}_{2}+\lambda^{i}_3\mathbf{f}_{k} \right)^{\top}\mathbf{1} = C.
\end{align*}
\vskip -4mm

\subsection{Actor Improvement using Standard Bellman}

The actor component improves a policy by computing $\R$ (Eq.~\ref{eq:rho_}) using the estimated Z-values from the critic because we do not assume knowledge of noise level  $\boldsymbol{\sigma}$. It is estimated by minimizing the least-square error between the V-value and the state-action Q-value:
\vskip -5mm
\begin{align*}
\min_{\R} \frac{1}{2}\int _{\x}\left( \hat{Q}(\x,\hat{\uu}(\x)) - V(\x) \right)^2 d\x,
\end{align*}
\vskip -3mm
where \trim{$M$ is the number of samples and a subscript $j$ is a sample index.} $V$ is the true V-value and $\hat{Q}$ is the estimated Q-value under the estimated action  $\hat{\uu}(\x)$. Notice from Eq.~\ref{eq:optimraw} that a value for $S$ results in a policy as $B$ is known. Thus, we seek the $S$ that yields the optimal policy by minimizing the error because the Q-value equals V-value iff $\hat{\uu}$ is maximizing.

Analogously to the critic, we minimize the least-square error given above, $\hat{Q}^{i}_k - V_k$, with TD-learning. To formulate the TD error for the standard Bellman update, let $\xkk$ be a sample at the next time step given state $\xk$ under passive dynamics, and let $\tilde{\x}_{k+1} \coloneqq \xkk + B\huk\dt$ be the next state using control dynamics. Rearranging terms of the Bellman update given in Eq.~\ref{eq:Bellman_rev}, the TD error $d^{i}_k$ is,
\vskip -5mm
 \begin{align*}
 d^{i}_k \coloneqq r(\xk,\huk) + \hat{V}^{i}(\tilde{\x}_{k+1}) -\hVavg -\Vkt^{i}.
 \end{align*}
 \vskip -3mm
 We may use Eqs.~\ref{eq:totalcost} and~\ref{eq:rho_} to replace the reward function,
\vskip -6mm
\begin{align*}
 r(\xk,\huk) &= (q_k + 0.5\huk^{\top}(\hR^{i})^{-1}\huk)\dt = \qk + 0.5\frac{\partial \hat{V}_k^{i}}{\partial \xk}^{\top}B\hR^{i}B^{\top}\frac{\partial \hat{V}_k^{i}}{\partial \xk}\dt.
\end{align*}
\vskip -3mm
The last step is obtained by noting that $\huk^{i} = -\hR^{i} B^{\top}\frac{\partial \hat{V}_k^{i}}{\partial \xk}$, where $\hR$ denotes estimated $\R$ in Eq. \ref{eq:optimraw}. The estimated V-value and its derivative is calculated by utilizing the approximate Z-value function from the critic. $\hR$ is updated based on standard stochastic gradient descent using the TD error,
\vskip -5mm
\begin{align*}
& \hR^{i+1} = \hR^{i} - \beta^i \frac{\partial}{\partial \hR^{i}}\left ( \hat{Q}^{i}_k - V_k \right)^2 \approx \hR^{i} - 2\beta^i d^{i}_k \frac{\partial d^{i}_k}{\partial \hR^{i}},
\end{align*}
\vskip -3mm
where $\beta$ is the learning rate. The actor mitigates the impact of error from estimated Z-value by minimizing the approximated least-square error between V- and Q-values under the learned policy.
\trim{
The gradient of the TD error $d_k^i$ is used for updating the weight $\hR$ next, $\hR^{i+1} = \hR^{i} - \beta^i d^{i}_k \nabla_{\hR^{i}}d^{i}_k$ where $\beta^{i}$ is a learning rate. As the final step, the improved policy is obtained by using this updated estimate $\hR^{i+1}$ and derivative of updated V-value in Eq.~\ref{eq:optimraw}.}

\subsection{Algorithm}
\label{sec:algo}
\vspace{-3mm}
\begin{algorithm}[!ht]
\small
\caption{passive Actor Critic}\label{alg:pAC}
\begin{algorithmic}[1]
\State {Initialize parameters $\hZavg^0, \paramz^0, \hR^0, \alpha_1^0, \beta^0$}
\For {Iteration $i=1$ to $N$}
\State {Sample set of a state, a next state and state cost $(\xk, \xkk, q_k)$ from dataset randomly}
\BState \emph{critic}:
\State {$ e^{i}_k  \gets \hZavg^{i}\Zkt^{i} - \exp{(-q_k)}\Zkkt^{i}$}
\State {Update $\paramz^{i+1}$ with $2e^{i}_k \hZavg^i\frac{\partial \Zkt^{i}}{\partial \paramz^{i}}$}
\State {$\hZavg^{i+1} \gets \hZavg^{i} - 2\alpha_1^{i} e^{i}_k\Zkt^{i}$}
\BState \emph{actor}:
\State {$\hat{V}_k^i \gets -\ln{\Zkt^i}$, $\hat{V}_{k+1}^i \gets -\ln{\Zkkt^i}$, $\hVavg^i \gets -\ln{\hZavg^i}$}
\State {$d^i_k \gets q_k + 0.5\frac{\partial \hat{V}_k^{i}}{\partial \xk}^{\top}B\hR^{i}B^{\top}\frac{\partial \hat{V}_k^{i}}{\partial \xk}\dt + \hat{V}_{k+1}^{i} -\hVavg^i -\Vkt^{i}$}
\State {$\hR^{i+1} \gets \hR^{i} - 2\beta^i d^{i}_k \frac{\partial d^{i}_k}{\partial \hR^{i}}$}
\EndFor
\end{algorithmic}
\end{algorithm}

We show a pseudo code of pAC in Algorithm~\ref{alg:pAC}. $Z(\x)$ and $\Zavg$ are estimated in the critic with samples, and $\R$ is done in the critic with samples, estimated $\hat{Z}(\x)$ and $\hZavg$. In the critic, feedback from the actor is not needed (unlike actor critic methods for MDPs) because the Z-value is approximated with samples from passive dynamics only. We emphasize that the actor and critic steps do not use the functions $A$ and $\boldsymbol{\sigma}$ but does indeed rely on $B$, of Eq.~\ref{eq:dynamics}. As such, the updates use a sample $({\mathbf x}_k,{\mathbf x}_{k+1})$ of the passive dynamics, and the state cost $q_k$. Consequently, pAC achieves semi model-free learning for L-MDPs.

\section{Numerical Experiments}
\label{sec:experiments}
We evaluate pAC for L-MDPs on two synthetic domains, Car-on-a-Hill and Pendulum, also used previously on L-MDPs by Todorov~\cite{todorov2009eigenfunction}; and on our motivating domain of autonomous merging in a congested freeway.

\subsection{Problem settings}

\begin{table}[!t]
\centering
\small
\begin{tabular}{|c|c|c|c|}
\hline
& Car-on-a-Hill&Pendulum&Simulated freeway merge\\
\hline
\hline
$A(\x)$ & $\begin{bmatrix}
                x_v (1+s(x_p))^{-\frac{1}{2}}  \\
                -9.8~{\rm sign}(x_p)\left ( 1+s(x_p)^{-2} \right)^{-\frac{1}{2}}
            \end{bmatrix}$ & $\begin{bmatrix}
                x_v,\,
                \sin(x_p)
            \end{bmatrix}^\top$ & $\begin{bmatrix}
                dv_{12},\,0,\,
                dv_{02}+0.5a_{0}({\mathbf x})\dt ,
                a_{0}({\mathbf x})
            \end{bmatrix}^\top$ \\
\hline
$B$ & $\begin{bmatrix} 0,\,1 \end{bmatrix}^\top$ & $\begin{bmatrix}
             0,\,1
            \end{bmatrix}^\top$& $\begin{bmatrix}
0.5\dt,\,1,\,0,\,0
\end{bmatrix}^\top$ \\
\hline
$\boldsymbol{\sigma}$ & $\begin{bmatrix}
             0,\,1
            \end{bmatrix}^\top$ & $\begin{bmatrix}
             0,\,2
            \end{bmatrix}^\top$ & $\begin{bmatrix}
             0,\,2.5,\,0,\,2.5
            \end{bmatrix}^\top$\\
\hline
\end{tabular}
\caption{\small System dynamics of each domains} \label{tb:dynamics}
\vspace{-5mm}
\end{table}

\noindent {\bf Car-on-a-Hill}~~Car-on-a-Hill has a two-dimensional state space, ${\mathbf x} = [x_p, x_v]^\top$ where $x_p$ and $x_v$ denote position and velocity, respectively, and a one-dimensional action space. Table \ref{tb:dynamics} shows the dynamics in detail.
\trim{ The dynamics are:
\begin{align*}
A({\mathbf x}) &=  \begin{bmatrix}
                x_v (1+s(x_p))^{-\frac{1}{2}}  \\
                -9.8~{\rm sign}(x_p)\left ( 1+s(x_p)^{-2} \right)^{-\frac{1}{2}}
            \end{bmatrix}, \\
\text{where } s(x_p) &=  0.5 x_p \exp{(-x_p^2)}\\
B &= \begin{bmatrix}
             0,\,1
            \end{bmatrix}^\top, \quad \boldsymbol{\sigma} = \begin{bmatrix}
             0,\,1
            \end{bmatrix}^\top, \quad \Delta t=0.01 [{\rm sec}].
\end{align*}
}
 The state cost is given by,
 \vskip -7mm
\begin{align*}
q({\mathbf x}) =& 4.0\left( \exp\left(-0.5(x_p-1)^2 -(x_v+1)^2 \right) + \exp\left(-0.5(x_p+1)^2 -(x_v-1)^2 \right) -2 \right).
\end{align*}
\vskip -4mm
Initial states are randomly set in $-2\pi \leq x_p \leq 2\pi$ and $-\pi \leq x_v \leq \pi$.\\
\vskip -3mm
\noindent {\bf Pendulum}~~The Pendulum problem also has a two-dimensional state space similar to the Car-on-a-Hill and a one-dimensional action space. Table \ref{tb:dynamics} shows the dynamics in detail.
\trim{ The dynamics are:
\begin{align*}
& A({\mathbf x}) =  \begin{bmatrix}
                x_v,\,
                \sin(x_p)
            \end{bmatrix}^\top,\\
& B = \begin{bmatrix}
             0,\,1
            \end{bmatrix}^\top, \quad \boldsymbol{\sigma} = \begin{bmatrix}
             0,\,2
            \end{bmatrix}^\top, \quad \Delta t=0.01 [{\rm sec}].
\end{align*}
}
State cost is given by
\vskip -7mm
\begin{align*}
q({\mathbf x}) =  4.0\left(\exp\left(-(x_v-3)^2 \right) +\exp\left(-(x_v+3)^2 \right) -2)\right).
\end{align*}
\vskip -4mm
Initial states are randomly set in $-2\pi \leq x_p \leq 2\pi$ and $-\pi \leq x_v \leq \pi$.\\

\begin{figure}[t]
\centering
\includegraphics[width=0.5\hsize, height=2.5cm]{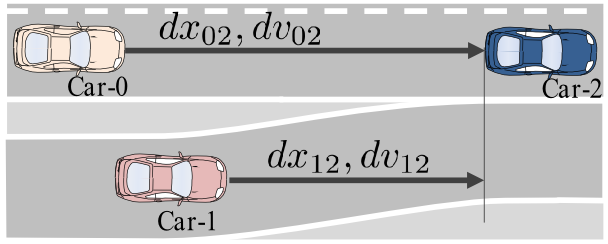}
\caption{\small The 3-car system for studying merging. Merging vehicle (Car-1) should ideally merge midway between the following vehicle (Car-0) and leading vehicle (Car-2). $dx_{12}$ and $dv_{12}$ denote Car-1's relative position and velocity from Car-2.}
\label{fig:A-B-C}
\vskip -3mm
\end{figure}
\vskip -3mm
\noindent {\bf Simulated freeway merging}~~This new contemporary domain simulates freeway merges by an automated vehicle. We refer the reader to Fig.~\ref{fig:A-B-C} for establishing the four-dimensional state space. Here, ${\mathbf x} = [dx_{12}, dv_{12}, dx_{02}, dv_{02}]^\top$ where $dx_{ij}$ and $dv_{ij}$ denote the horizontal signed distance and relative velocity between cars $i$ and $j\in [0,1,2]$. The action space is one-dimensional (acceleration). Table \ref{tb:dynamics} shows the dynamics in detail.
\trim{ The dynamics are:
\begin{align*}
A({\mathbf x}) &=  \begin{bmatrix}
                dv_{12},\,0,\,
                dv_{02}+0.5a_{0}({\mathbf x})\dt\,
                a_{0}({\mathbf x})
            \end{bmatrix}^\top,\\
B &=  \begin{bmatrix}
0.5\dt,\,1,\,0,\,0
\end{bmatrix}^\top,\,
\boldsymbol{\sigma} = \begin{bmatrix}
             0,\,2.5,\,0,\,2.5
            \end{bmatrix}^\top,\\
a_{0}({\mathbf x}) &= \alpha \frac{v^{\beta}_{2}(-dv_{02})}{-dx^\gamma_{02}},\, \Delta t = 0.1 [{\rm sec}].
\end{align*}
}
The dynamics presume that the leading vehicle is driven with a constant speed $v_2=30$[m/sec], and the following vehicle is driven by a known car-following model~\cite{olstam960comparison}. The acceleration of the vehicle $a_0(\x)$ is calculated with $ -\alpha v^{\beta}_{2}dv_{02}/(-dx)^\gamma_{02}$, where if the following vehicle is slower than the leading vehicle ($dv_{02}<0$), $\alpha=1.55$, $\beta=1.08$, $\gamma$ = 1.65, otherwise $\alpha=2.15$, $\beta=-1.65$, $\gamma=-0.89$. The assumptions are used to simulate maneuvers of ambient vehicles in only the simulated freeway merge domain.

The state cost designed to motivate Car-1 to merge midway between Cars 0 and 2 with the same velocity as Car-0, is:
\vskip -7mm
\begin{align*}
q({\mathbf x}) &= k_1  - k_1\exp \left( -k_2 \left(1-\frac{2dx_{12}}{dx_{02}}\right)^2 - k_3 dv_{10}^2\right)
\end{align*}
where $k_1$, $k_2$ and $k_3$ are weights for the state cost ($[k_1,k_2,k_3]=[1,10,10] {~~\rm if~~} dx_{02}<dx_{12}<0,~ [10,10,0] {~~\rm otherwise} $). Initial states are randomly picked in $-100 < dx_{12} < 100$ [m], $-10 < dv_{12} < 10$ [m/sec], $-100 < dx_{02} < -5$ [m], and $-10 < dx_{02} < 10$ [m/sec].

\subsection{Performance Evaluation}
\begin{figure*}[!t]
\centering
	\begin{tabular}{ccc}
    	\begin{minipage}[t]{0.33\hsize}
        	\includegraphics[width=1.0\hsize]{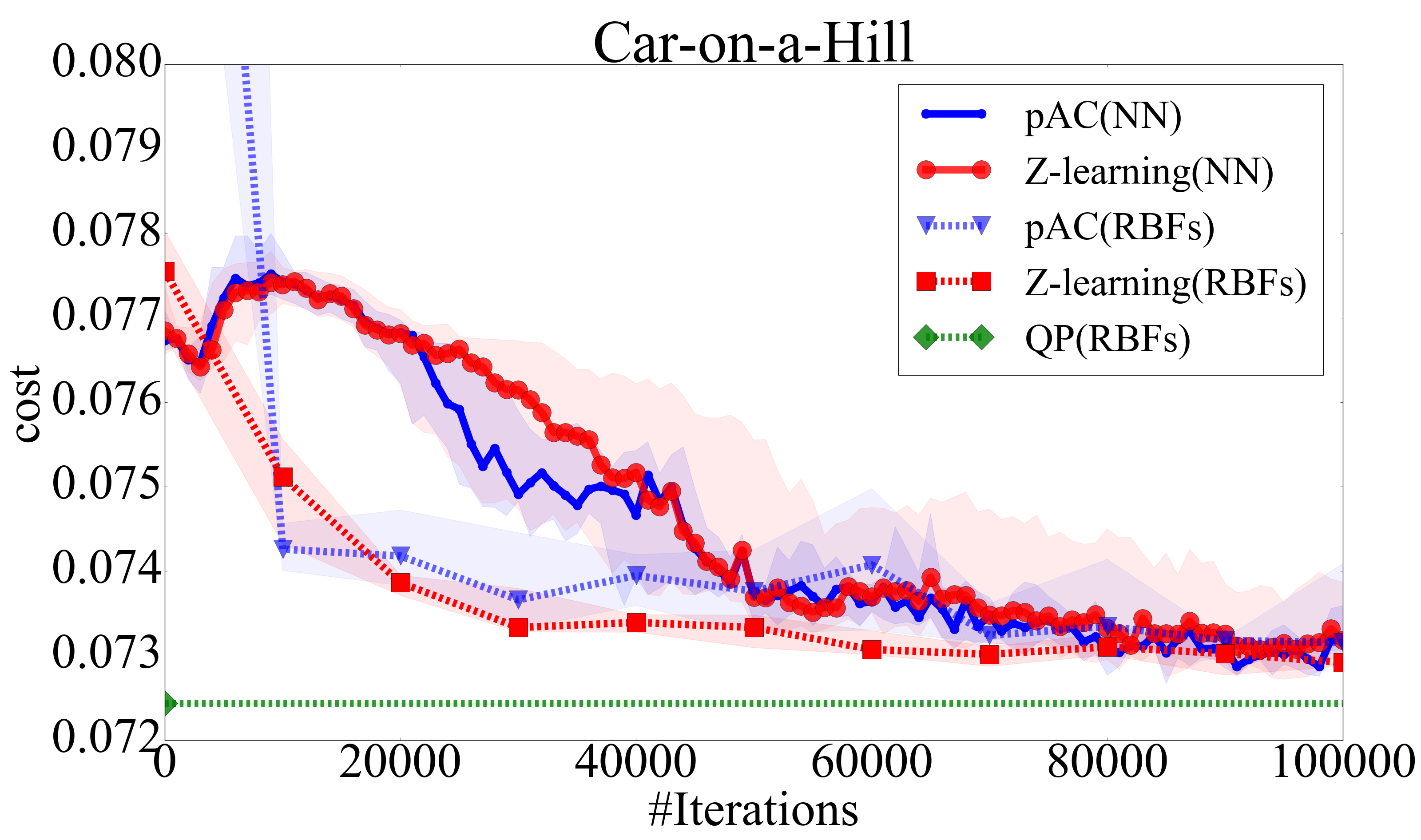}
    	\end{minipage}
    	\begin{minipage}[t]{0.33\hsize}
        	\includegraphics[width=1.0\hsize]{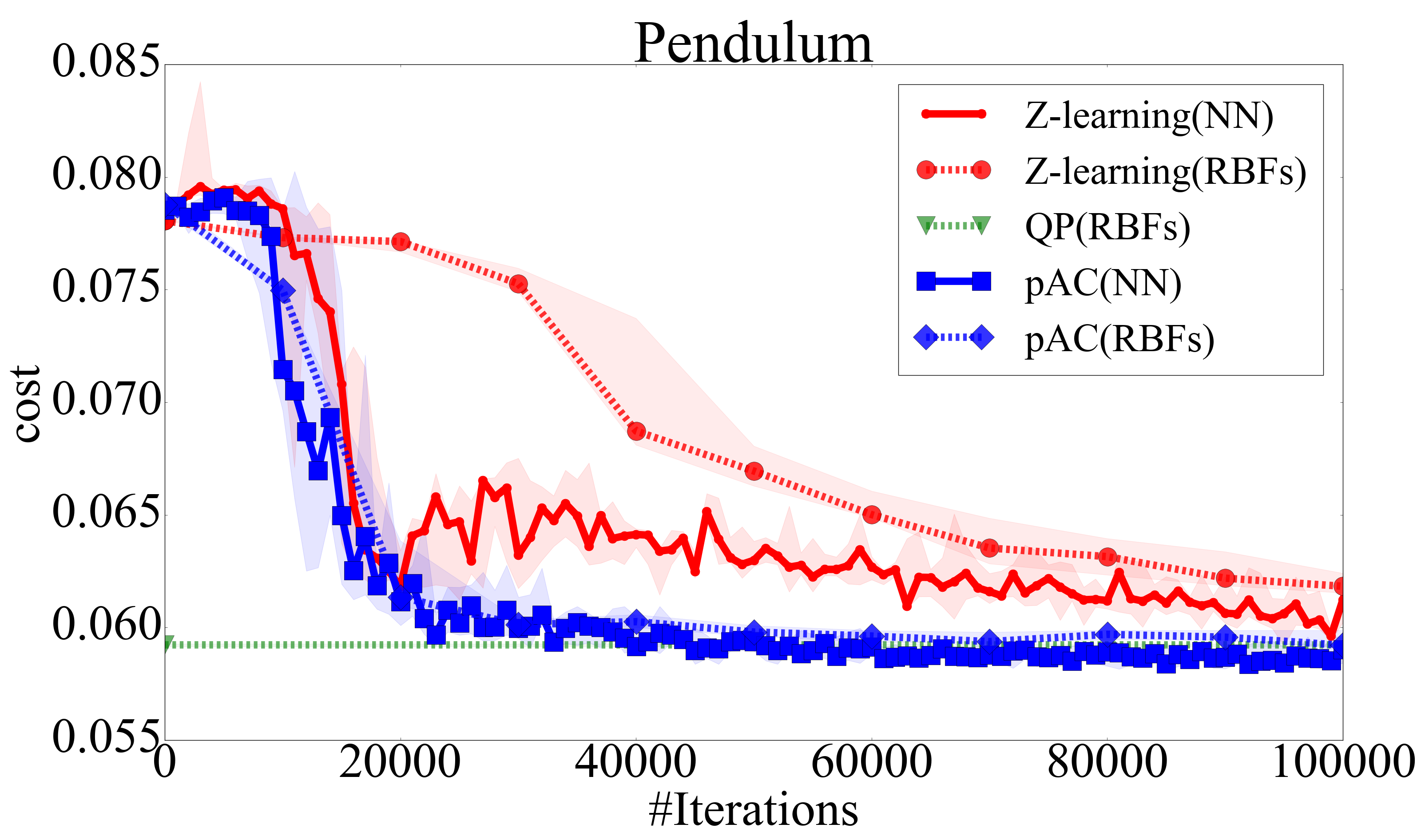}
    	\end{minipage}
    	\begin{minipage}[t]{0.33\hsize}
        	\includegraphics[width=1.0\hsize]{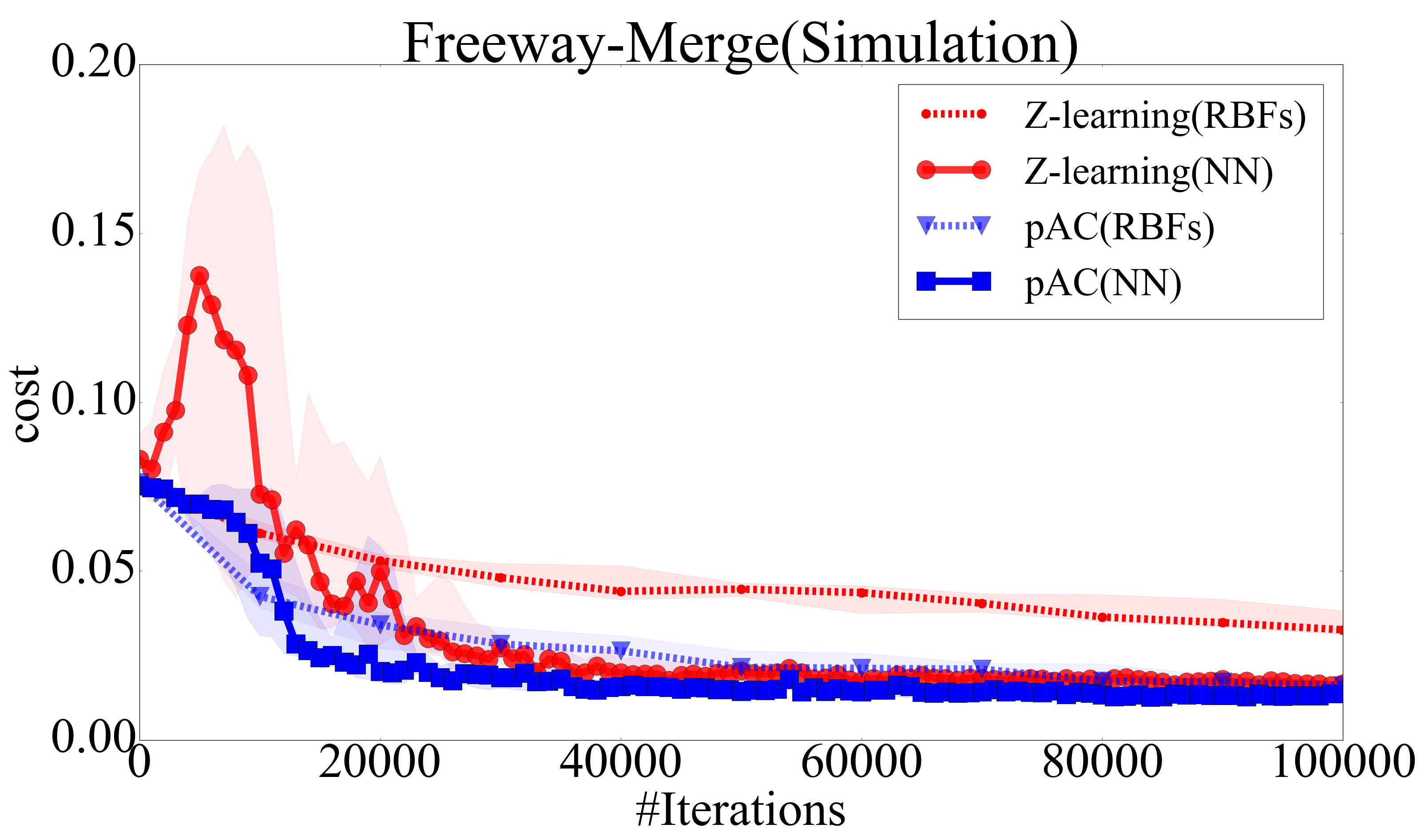}
    	\end{minipage}
	\end{tabular}
	\vskip -2mm
	\caption{\small Learning curves of average costs for pAC(blue), QP(green) and Z-learning(red) in three experiments. Results of pAC and Z-learning are illustrated in Freeway-Merge as QP could not learn any reasonable policies. }\label{fig:result-avecost}
\end{figure*}
We compared pAC with two other methods\trim{\footnote{Another method for a finite horizon formulation~\cite{uchibe2014combining} was not included in the evaluations as it is unclear how to compare it with others with infinite horizon.}}: model-based learning based on quadratic programming (QP) and Z-learning. QP requires all system dynamics and approximates the Z-value with quadratic programming~\cite{todorov2009efficient}. Z-learning assumes that $B$ and $\boldsymbol{\sigma}$ are available to approximate Z-value using the critic. QP and Z-learning calculate the policy with Eq. \ref{eq:optimraw}. We are unaware of any fully model-free method for L-MDPs. model-free PI control (e.g. \cite{Theodorou2010}) assume action cost is available and it is equivalent to the assumption of the known transition noise level in L-MDPs. Table~\ref{tb:comparisons} gives the prior knowledge requirement on  components of the system dynamics in Eq. \ref{eq:dynamics}.

\begin{table}[!t]
\centering
\begin{tabular}{|c|c|c|c|}
\hline
&$A({\mathbf x})$&$B$&$\boldsymbol{\sigma}$\\
\hline
\hline
QP & Known & Known & Known \\
\hline
Z-learning & \bf{Unknown} & Known & Known \\
\hline
pAC & \bf{Unknown} & Known & \bf{Unknown} \\
\hline
\end{tabular}
\caption{\small Model requirements for the evaluated methods. Notice that pAC imposes least requirement on prior model knowledge.} \label{tb:comparisons}
\vspace{-8mm}
\end{table}

\begin{figure}[!ht]
\centering
	\begin{tabular}{cc}
        \begin{minipage}[t]{0.45\hsize}
            \includegraphics[width=1.0\hsize]{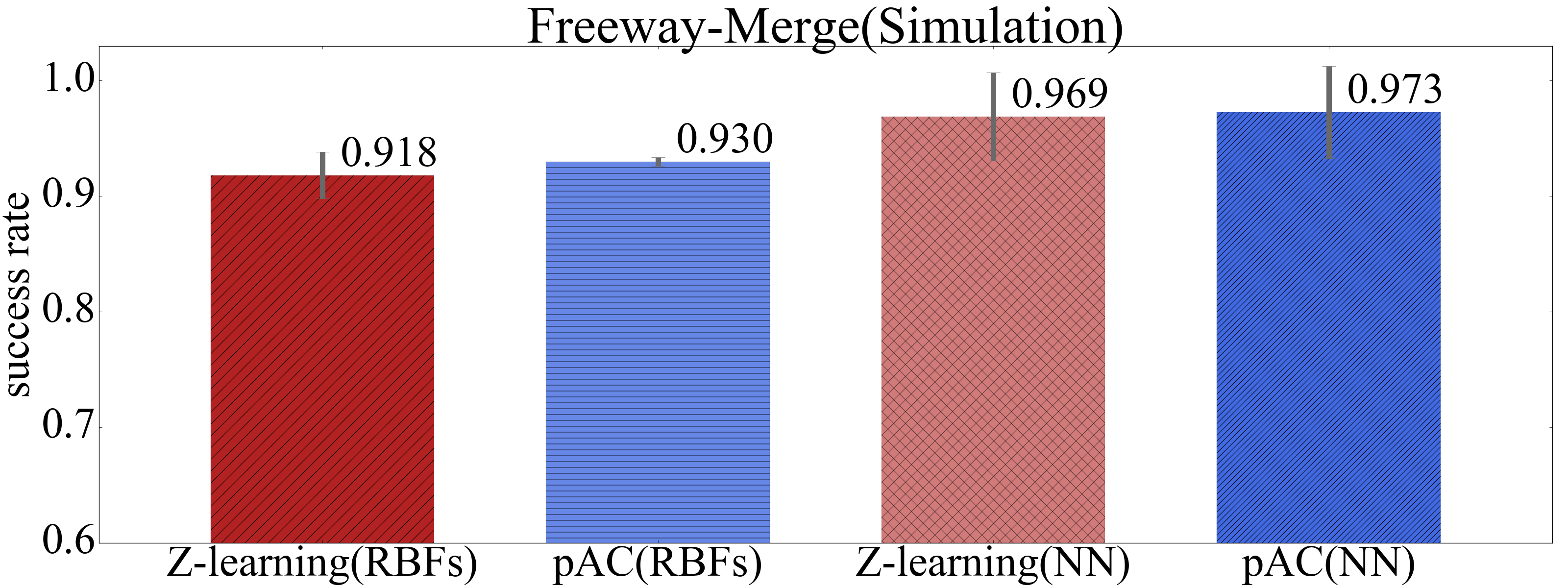}
        \end{minipage}
        \begin{minipage}[t]{0.45\hsize}
            \includegraphics[width=1.0\hsize]{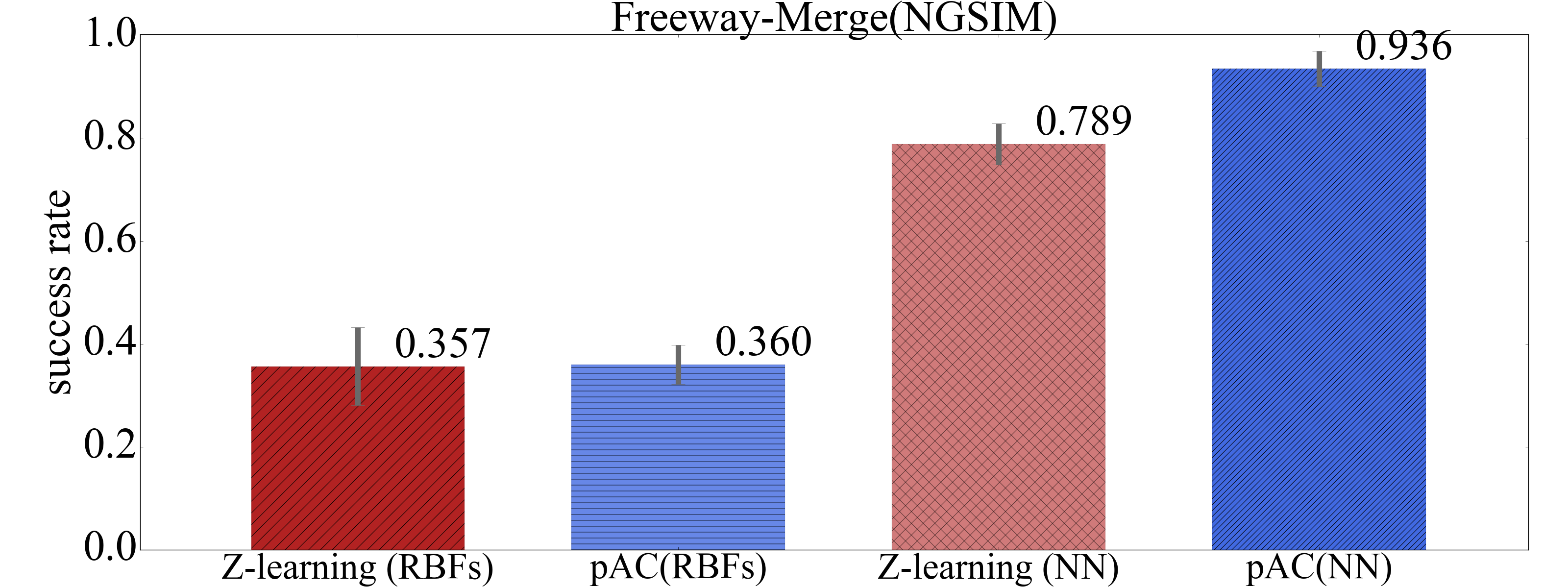}
        \end{minipage}
    \end{tabular}
    	\vskip -2mm
		\caption{\small (Left) Merging success rate for pAC and Z-learning in 30 seconds. RBFs and NN were the two function approximators used. (Right) Merging success rate for pAC and Z-learning on NGSIM. $\R$ is estimated by a Gaussian process for Z-learning as true dynamics are not available. Raw and ANN denote states sampled randomly and with ANN from data.} 	\label{fig:success-rate-freewaymerge}
\end{figure}

Gaussian RBFs are used to approximate the Z-value function in all methods and NNs are additionally used in Z-learning and pAC. RBFs were spaced uniformly in the range of sampled data: 400 RBFs were used for Car-on-a-Hill and Pendulum domains, and 4,096 RBFs were used for the Merging task. The standard deviations of the bases were 0.7 of the distance between the closest two bases in each dimension.

A three-hidden layer perceptron with 200, 200 and 50 units in first, second and third hidden layers is used as the NN. The number of nodes in the input is same as the dimensions of state space and one output, respectively. The rectified linear function~\cite{nair2010rectified} is used as the hidden layers' activation function. The activation function in the output layer for Car-on-a-Hill is $\exp(-\rm{tanh}(x))$ and $\exp(-\rm{softplus}(x))$ for other domains. We estimated $\R$ as constant in pAC. Inputs of the perceptron were normalized to the range $[0,1]$.

In Fig. \ref{fig:result-avecost}, we compared average cost calculated over periods of 10 seconds for Car-on-a-Hill and Pendulum, and 30 seconds for Merging, under learned policies in each domain. Observe that pAC finds a similar or better policy in comparison to other learning methods in all domains. QP could not learn any reasonable policy in Merging domains because of an issue referred to in ~\cite{todorov2009eigenfunction}: QP might not converge to a principal Eigen pair, instead of converging  to a 2nd or higher-order pair.

This improvement over Z-learning may come as a surprise because Z-learning makes greater use of model knowledge -- it calculates $\R$ using the true $B$ and $\boldsymbol{\sigma}$. Nevertheless, the presence of the additional {\em actor} step in pAC makes the difference. The actor additionally minimizes the error in Q-value to obtain $\R$. This error is not minimized if $\R$ is obtained as in Eq.~\ref{eq:rho_} due to the approximation error of Z-value. Subsequently, pAC estimates $\R$ differently, in a more targeted way to obtain a better policy.

\trim{We also find that function approximation with NN improves the performance over RBFs for both pAC and Z-learning. NNs are more effective for Z-learning than for pAC because the function approximation error more directly impacts the performance of Z-learning as pAC mitigates its effect by the actor. This result is significant as it indicates that pAC could be more robust to various function approximators.}

We evaluated the rate of merging successfully in a time limit of 30 seconds starting from 125 different states in Freeway-Merge. We defined success as being between the leading and following vehicles after 30 seconds. Figure~\ref{fig:success-rate-freewaymerge} (a) shows the success rate: pAC with RBFs and NN achieved $93\%$ and $97\%$ success rate respectively, which is comparable to Z-learning. All results shows pAC is comparable or better than Z-learning despite less prior knowledge instead of using actor step.

\trim{Of course, any model-free RL method would also learn proper policies in the above domains without prior knowledge of the dynamics. However, active exploration is required and would be unacceptable in some domains such as automated driving. The objective of the evaluations was to confirm that our proposed method could also learn compatible policy like such prior method without the active exploration and the passive dynamics model.}
\subsection{Experiment on real-world traffic}
\label{sec:ngsim}

The NGSIM data set contains vehicle trajectory data recorded by cameras mounted on top of a building \trim{on eastbound Interstate-80 in the San Francisco Bay area in Emeryville, CA} for 45 minutes around the evening rush hour\cite{ngsim}. Vehicle trajectories were extracted using a vehicle tracking method from collected videos \cite{kovvali2007video}. We extracted three-vehicle systems (Fig.~\ref{fig:A-B-C}) representing 637 freeway merge events.\trim{and applied a Kalman smoother to mitigate vehicle tracking noise.}

We compared pAC and Z-learning based on RBFs and NN on the extracted data from NGSIM. Z-learning calculated a policy with transition noise $\boldsymbol{\sigma}$ estimated with a Gaussian process due to unknown true dynamics. We used the same state variables, action variable and reward function as used in the simulated Freeway-Merge domain. We calculated next states $\xkk$ under passive dynamics by subtracting state change caused by actions:$\xkk = \x^{D}_{k+1} - B^{\top}\uk^{D}$, where $\x^{D}_{k+1}$ and $\uk^{D}$ are the next state and the action recorded in the data set respectively. We resampled data to mitigate any imbalance and sparseness by sampling randomly a state and choosing a nearest data from the state using the approximate nearest neighbor method.

\begin{figure}[!ht]
\centering
		\includegraphics[width=0.6\hsize]{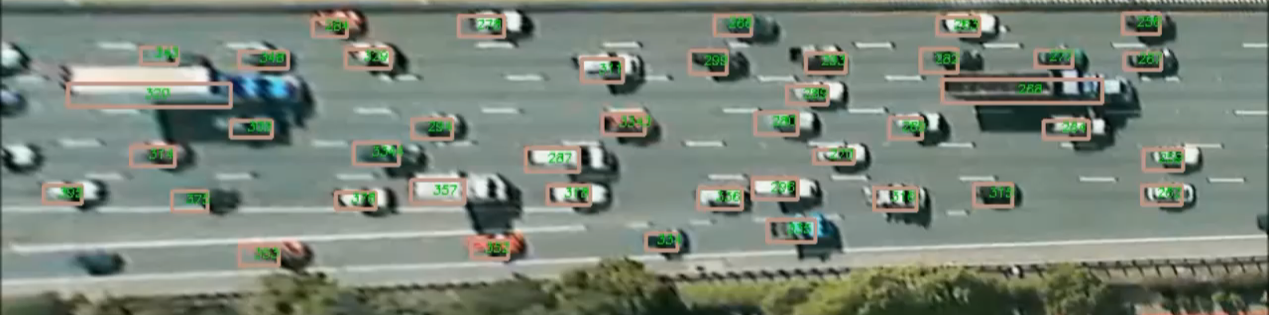}
		\caption{\small A snapshot of vehicles tracked in NGSIM. The bounding boxes denote the positions of tracked vehicle and numbers are vehicle indices.}\label{fig:ngsim-dataset}
\end{figure}

Success was defined as being between Car-0 and Car-2 at the merging point, for those instances in the data set where Car-1 on entry ramp completed its merge. Trajectories of Car-0 and Car-2 were played back from recorded logs, and trajectories of Car-1 were simulated with the control dynamics and learned policies. The dataset was randomly partitioned into five sub-datasets where an almost equal number of trajectories were included. Four sub-dataset were used as training data and a remaining sub-dataset were used as the test data for testing the policy. Each method was evaluated five times for each different test data.
\trim{
\begin{figure}[!ht]
\centering
	    \includegraphics[width=0.8\hsize]{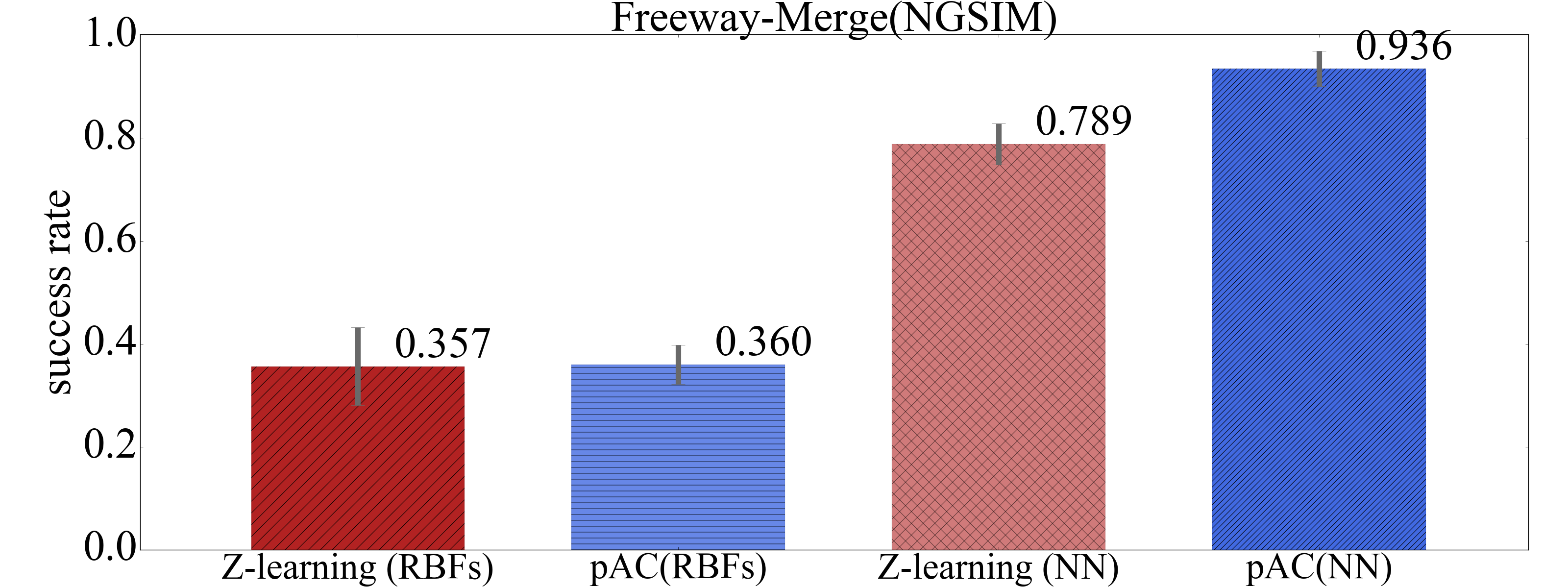}
		\caption{\small Merging success rate for pAC and Z-learning on NGSIM. $\R$ is estimated by a Gaussian process for Z-learning as true dynamics are not available. Raw and ANN denote states sampled randomly and with ANN from data.}\label{fig:success-rate-freewaymergeNGSIM}
\end{figure}
}
Figure~\ref{fig:success-rate-freewaymerge}(b) shows the success rate for each method. Firstly, we note the significant performance improvement when NN is used as the function approximator. Both pAC and Z-learning with RBFs simply do not perform well. Specifically, pAC with NN achieved a 93\% success rate significantly outperforming Z-learning with NN. The result shows that the actor, which minimizes the error of Q-value, improves performance more than the noise level estimation from the dataset. The model-free RL approaches cannot be applied here due to the need for active exploration.

\section{Concluding Remarks}
\label{sec:conclusion}

We presented a novel method for semi model-free RL for L-MDPs -- an important subclass of MDPs. The passive actor-critic optimizes a policy without active exploration, instead of using samples of the passive dynamics and knowledge of the control dynamics. This is a first formulation of the actor-critic schema in the context of L-MDPs. We evaluated the method using three domains. Results show that pAC achieves comparable or better performances than benchmarked methods despite less prior knowledge requirements. As such, pAC represents a significant step toward more efficient RL in continuous domains.

Data under passive dynamics and accurate models of control dynamics are needed for pAC. This may seem to limit applicability apparently but, as mentioned, it is well suited for contemporary robotic applications such as automated driving in real-world traffic. In this case, passive and control dynamics correspond to models of ambient and autonomous vehicles, respectively. pAC obtains a better policy with the car's own dynamics model, which is now commonly available, and uses data collected on maneuvers of the ambient vehicles whose models are usually not known. An evaluation of the learned policies toward freeway merging using a real traffic data set illustrates its usefulness for practical applications. We are interested in exploring additional challenges, e.g., entering roundabouts.


\bibliographystyle{plain}

\end{document}